%% file: root.tex
\documentclass[letterpaper, 10 pt, conference]{ieeeconf}  

\IEEEoverridecommandlockouts                              

\input{shortcuts.tex}

\title{\LARGE \bf
Vision-Based Mobile Robotics Obstacle Avoidance With Deep Reinforcement Learning
}

\author{Patrick Wenzel\textsuperscript{1} \ \ \
Torsten Schön\textsuperscript{2} \ \ \
Laura Leal-Taixé\textsuperscript{1} \ \ \
Daniel Cremers\textsuperscript{1}
\thanks{\textsuperscript{1} P. Wenzel, L. Leal-Taixé, and D. Cremers are with the Technical University of Munich, Germany. \texttt{\{patrick.wenzel, leal.taixe, cremers\}@tum.de}}
\thanks{\textsuperscript{2} T. Schön is with the Technische Hochschule Ingolstadt, Ingolstadt, Germany. \texttt{torsten.schoen@thi.de}}
}

\begin{document}
\maketitle
\thispagestyle{empty}
\pagestyle{empty}


\input{sections/abstract.tex}



\input{sections/introduction}

\input{sections/background}

\input{sections/method}

\input{sections/experiments}

\input{sections/conclusion}

\bibliographystyle{IEEEtran}
\bibliography{library.bib}  

\end{document}

%% file: shortcuts.tex
\usepackage{amsmath,amssymb,amsfonts}
\usepackage{siunitx}
\usepackage{algorithmic}
\usepackage{graphicx}
\usepackage{textcomp}
\usepackage{pgfplots}
\pgfplotsset{compat=newest}
\usepackage{comment}
\usepackage{tikz}
\usepackage{subcaption}
\usepackage{authblk}
\usepackage{booktabs}
\usepackage{acronym}

\DeclareMathOperator*{\argmax}{argmax}

\acrodef{drl}[DRL]{deep reinforcement learning}
\acrodef{slam}[SLAM]{simultaneous localization and mapping}
\acrodef{cnn}[CNN]{convolutional neural network}
\acrodef{gan}[GAN]{generative adversarial network}
\acrodef{rl}[RL]{reinforcement learning}
\acrodef{dnn}[DNN]{deep neural network}
\acrodef{fov}[FOV]{field of view}

%% file: sections/abstract.tex
\begin{abstract}
Obstacle avoidance is a fundamental and challenging problem for autonomous navigation of mobile robots. In this paper, we consider the problem of obstacle avoidance in simple 3D environments where the robot has to solely rely on a single monocular camera. In particular, we are interested in solving this problem without relying on localization, mapping, or planning techniques. Most of the existing work consider obstacle avoidance as two separate problems, namely obstacle detection, and control. Inspired by the recent advantages of deep reinforcement learning in Atari games and understanding highly complex situations in Go, we tackle the obstacle avoidance problem as a data-driven end-to-end deep learning approach. Our approach takes raw images as input and generates control commands as output. We show that discrete action spaces are outperforming continuous control commands in terms of expected average reward in maze-like environments. Furthermore, we show how to accelerate the learning and increase the robustness of the policy by incorporating predicted depth maps by a generative adversarial network.
\end{abstract}

%% file: sections/introduction.tex
\section{Introduction}\label{sec:introduction}

Safe and effective exploration in an unknown environment is a fundamental and challenging problem for mobile robots. To develop an approach for addressing these challenges a robotic system is faced with enormous challenges in perception, control, and localization. Obstacle avoidance is indeed an important and crucial part of being able to successfully navigate. Typically, in robotic navigation problems, \ac{slam}~\cite{Engel2014,engel2017direct,mur2015orb} algorithms are a fundamental part of such a system. These algorithms tackle this problem by constantly updating a map of the environment while simultaneously keeping track of the robot's location. However, this is a challenging problem because it is hard to design features or mapping techniques that work under a wide variety of environments the robot may encounter. Therefore, we are interested in utilizing learning techniques for mapping sensor inputs to control commands. In particular, we are interested in solving this issue end-to-end in a data-driven way without relying on an obstacle map. However, despite all that outstanding success in a lot of computer vision problems, many recent deep learning approaches still have evident drawbacks for robotics~\cite{Wong2016}. 

\begin{figure}[t]
\centering
\begin{tikzpicture}[      
        every node/.style={anchor=south west,inner sep=0pt},
        x=1mm, y=1mm,
      ]   
     \node (fig1) at (0,0)
       {\includegraphics[width=\linewidth]{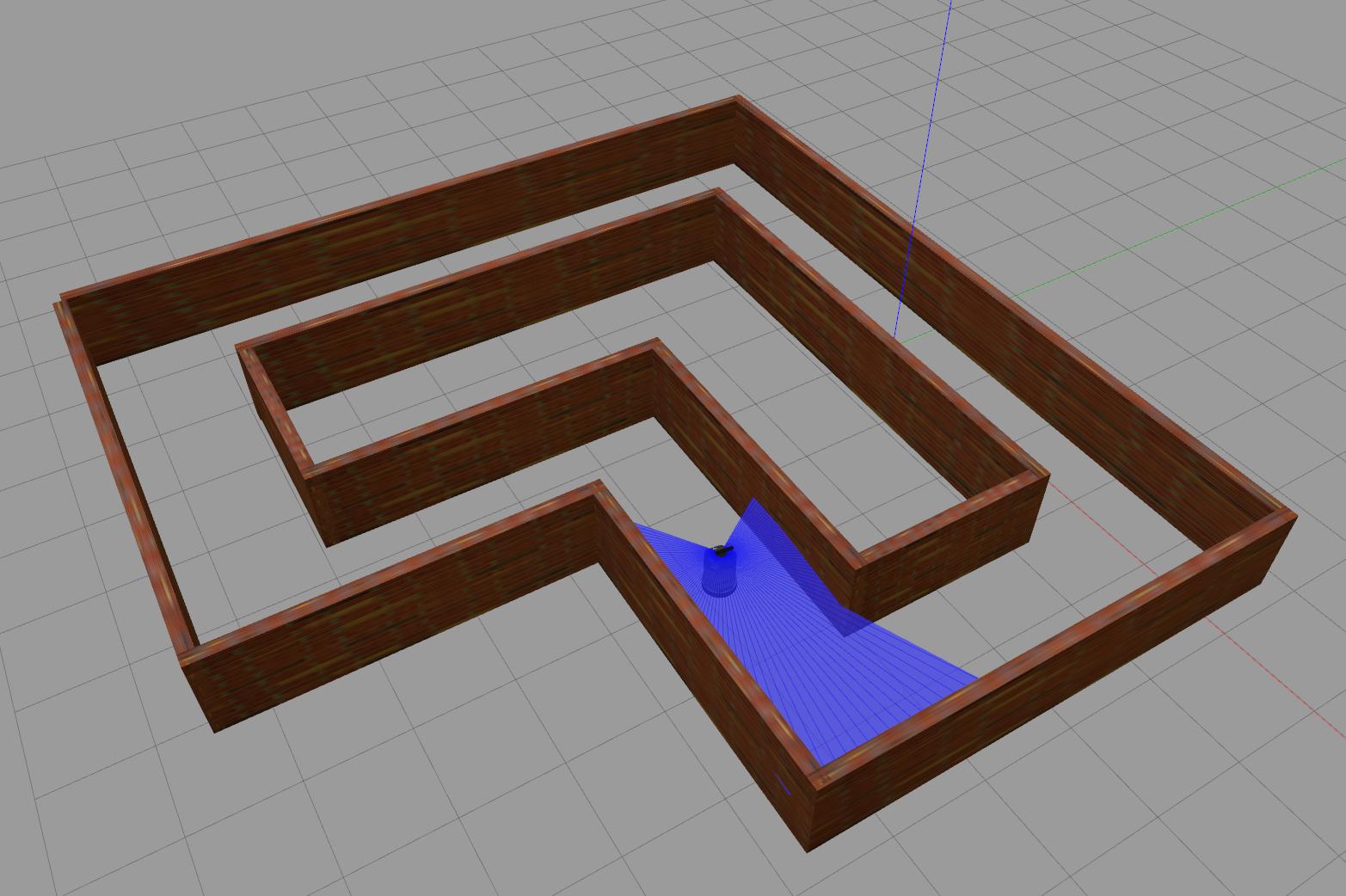}};
     \node (fig2) at (66,37)
       {\fbox{\includegraphics[scale=0.1]{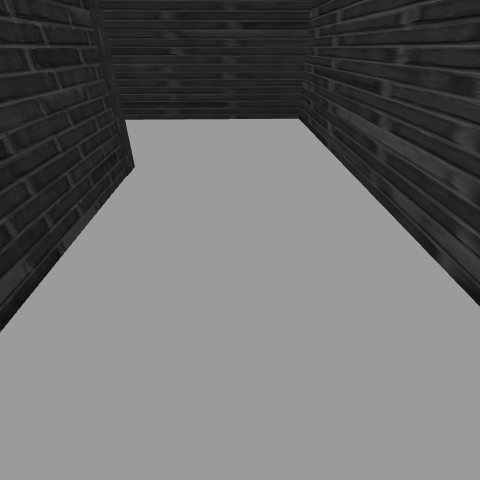}}};  
\end{tikzpicture}
\caption{A simulated mobile robot in a virtual environment learning how to navigate autonomously without crashing. The robot was trained based on grayscale images (example shown in the upper right part of the figure) captured by a monocular camera. The range finder readings are shown as blue lines emerging from the center of the agent.}
\label{fig:intro_figure}
\end{figure}

In the last few years, \ac{drl} has shown great potential to solve difficult problems that previously seemed beyond our grasp. While the improvement on tasks like Atari games~\cite{Mnih2015} and the game of Go~\cite{Silver2016} have been dramatic, the progress has been primarily in 2D environments. Therefore, in this paper, we are interested in the problem of obstacle avoidance in simple 3D environments for mobile robots without any hand-crafted features or prior knowledge of the environment (see Figure~\ref{fig:intro_figure}). We use \ac{drl} to learn visual control strategies that allow robots to explore an unknown environment by using raw sensor data only. Furthermore, we are particularly interested in the comparison between discrete and continuous action spaces for learning control strategies.

Deep learning approaches have shown huge success in image-based pattern recognition problems due to their ability to learn complex and hierarchically abstract feature representations. In this paper, we train visualmotor policies that perform both perception and control together for robotic navigation tasks. The policies are represented by deep \acp{cnn} which are being learned from raw visual input data, captured by a single monocular camera mounted on a mobile robot platform. For a more comprehensive overview of deep learning approaches in mobile robotics, we refer the reader to the survey by Tai and Liu~\cite{Tai2016}.

Since transferring knowledge learned in simulation to real-world settings is an important step in developing a complete robotics system, we are also interested in bridging the gap between those two. This is a rather challenging problem for perception-based approaches since the visual fidelity of images generated in simulation environments is inferior compared to images from the real world. This motivates us to tackle the problem by using depth images instead of images from a monocular camera. However, in practical scenarios perception is often limited to monocular vision systems without any access to depth images. Hence, we propose to use conditional \acp{gan}~\cite{Isola2017} to overcome this issue by predicting depth images from monocular images.

Particularly, this paper presents the following contributions:

\begin{itemize}
    \item We show that discrete action spaces outperform continuous action spaces on the task of vision-based robot navigation in maze-like environments.
    \item We investigate the influence of on-policy and off-policy \ac{drl} algorithms as well as the impact of different sensor modalities for visual navigation tasks in synthetic environments. The experimental findings can be used as a guideline for similar tasks. 
	\item We show that the robustness of the policy and the learning performance can be increased by additionally fusing predicted depth images generated by a generative model.
\end{itemize}

The remainder of this paper is organized as follows. In Section~\ref{sec:background}, the necessary background on \ac{drl} for robot navigation is reviewed. Section~\ref{sec:method} describes the problem statement and the approach. Training and evaluation details are presented in Section~\ref{sec:experiments}. Finally, Section~\ref{sec:conclusion} concludes the paper.

%% file: sections/background.tex
\section{Background}\label{sec:background}

In this work, we consider a Markov decision process (MDP), where an agent interacts with the environment through a sequence of observations, actions, and reward signals. At each time step $t$, the agent executes an action $a_t \in \mathcal{A}$ from its current state $s_t \in \mathcal{S}$, according to its policy $\pi \colon \mathcal{S} \to \mathcal{A}$. The received reward at time $t$ which is obtained after interaction with the environment is denoted by $r_t \colon \mathcal{S} \times \mathcal{A} \to \mathbb{R}$ and transits to the next state $s_{t+1}$ according to the transition probabilities of the environment. The aim of the agent is to find a policy $\pi(a{\mid}s;\theta)$ (where $\theta$ are the parameters of the policy) that maximizes the sum of discounted rewards, $R_t = \sum_{\tau = 0}^{T} \gamma^{\tau} r_{t + \tau}$, in expectation, i.e., $\mathbb{E}[R_t]$, where $\gamma$ is the discount factor and $T$ is the time horizon. The discount factor trades off the balance between the immediate and the future reward. Since the transition probabilities are unknown, \ac{rl} techniques can be used to learn a policy for navigating a mobile robot in an unknown environment without crashing.

\subsection{Deep Reinforcement Learning}
In a model-free \ac{rl} setting, the agent is trained without access to the underlying model of the environment. However, \ac{drl} methods exploit the idea of using \acp{dnn} to approximate the value function, the policy, or even the model. In general, today's leading \ac{drl} algorithms can be divided into two main classes: off-policy methods and on-policy methods. In the first method, the policy used for the control, called the behavior policy, may have no correlation to the policy that is being updated, called the estimation policy. In the second method, the policy used for the control of the MDP is the same which is being updated. In the following, we briefly introduce two recent model-free \ac{rl} methods.

\textit{1) Deep Q-Network:} Deep Q-Network (DQN) proposed by Mnih et al.~\cite{Mnih2015} is an off-policy method and the action-value function $Q(s, a; \theta)$ is learned by minimizing the temporal difference error instead of directly parameterizing a policy. The Q-value estimator is optimized by repeated gradient descent steps on the objective: $\mathbb{E} \left[ (y_i - Q(s_i, a_i; \theta))^{2} \right]$, where $y_i$ is the estimated Q-value given by $y_i = r_i + \gamma \max_{a} Q(s_{i+1}, a; \theta)$. The policy selects the action maximizing value: $a^{*} = \argmax_a Q(s, a; \theta)$. We make use of several enhancements to DQN as suggested by~\cite{Hessel2017}, like dueling networks~\cite{Wang2016}, double Q-learning~\cite{VanHasselt2015} and prioritized experience replay~\cite{Schaul2015} to improve performance. We use the proposed hyperparameters by OpenAI baselines~\cite{baselines} except a reduced experience replay capacity of $\num{2.5e5}$ samples. 

\textit{2) Proximal Policy Optimization:} Proximal Policy Optimization (PPO)~\cite{Schulman2017} is a recently released on-policy method that improves sample complexity compared to standard policy gradient methods. In policy gradient methods, the policy is directly parameterized in the form $\pi(a{\mid}s;\theta)$, where $\pi$ is a probability distribution over actions $a$ when observing state $s$, as parameterized by $\theta$, a deep neural network. The objective to optimize is the following:
\begin{align*}
    \hat{E}_t \left[\min(r_t(\theta) \hat{A}_t, \operatorname{clip}(r_t(\theta), 1-\epsilon, 1+\epsilon) \hat{A}_t)\right],
\end{align*}
where $\hat{A}_t$ is the estimated advantage, $\hat{E}_t$ is the empirical expectation over timesteps, and $\epsilon$ a hyperparameter.

\subsection{DRL for Navigation}
Perception and control for robotics with \ac{drl} have been studied recently in the context of obstacle avoidance and planning for mobile robots. A mapless motion planner, using laser range findings, previous velocities and relative target positions as sensory input trained with Asynchronous Deep Deterministic Policy Gradient (ADDPG) was successfully developed by~\cite{Tai2017b}. The experiments show that the proposed method can navigate the mobile robot in virtual and real environments to the desired targets without colliding with any obstacles. Mirowski et al.~\cite{Mirowski2016} proposed to jointly learn the goal-driven reinforcement learning problem with auxiliary depth prediction and loop closure classification tasks. The results show that the combination with supervised auxiliary tasks significantly enhances the overall learning performance. Also,~\cite{Tai2016a} demonstrated the ability to build a cognitive exploration strategy using an RGB-D sensor and DQN. Furthermore,~\cite{Tai2016a} proved the ability to transfer the models trained solely in virtual environments to the real world. The navigation policies generalized well to the previously unseen real environments with dynamic obstacles. Compared to previous work, we show how to accelerate the learning and how to increase the robustness of the policy by means of fusing information from two different sensor modalities. Moreover, in this work we suggest guidelines on which algorithms and sensor modalities are adequate for visual navigation tasks and how discrete or continuous action spaces influence the performance. 

Another line of research is to use evolutionary methods for visual reinforcement learning tasks. In one example, Koutník et al.~\cite{Koutnik2013} proposed the idea of using evolutionary computation to train vision-based control policies for the TORCS driving game.

%% file: sections/method.tex
\section{Deep Reinforcement Learning for Control}\label{sec:method}
In this paper, we focus on incorporating predicted depth information into off-policy and on-policy methods for visual control of a mobile robot based on raw sensory data. Our vision-based control problem can be considered as a decision-making task, where the agent is interacting in an unknown environment with its sensors. 

\subsection{Reward Function}
Reward shaping is arguably one of the biggest challenges for every \ac{rl} algorithm. One big drawback of \ac{rl} is the fact that reward functions are mostly hand-engineered and domain-specific. There is quite some work on imitation learning (IL)~\cite{Ho2016}, which tries to directly recover the expert policy and inverse reinforcement learning (IRL)~\cite{Fu2017a}, which provides a way to automatically acquire a cost function from expert demonstrations. However, most of these methods suffer from heavy computation costs and the optimization of finding a reward function that best represents the expert trajectory is essentially ill-posed~\cite{Arora2018}.

There are two approaches on how to design a reward function. Dense reward functions, giving reward in all states during exploration and in contrast, sparse rewards, which only provide a reward at the terminal state, and no reward anywhere else. For the obstacle avoidance problem, a dense reward is much easier to learn, since the reward function will provide feedback about every step even when the policy hasn't figured out a full solution to the problem yet. Whereas a sparse reward function won't allow for safe navigation of the environment and also crashes will be more likely, e.g., since we don't punish the robot for navigating close to the walls. We define the dense reward at timestep $t$ as follows:

\begin{align*}
    r_t &= 
        \begin{cases}
         r_{\text{collision}}, & \text{if collides,} \\
         \left(\frac{1}{c_d + 1}\right)^2 + \frac{1}{2} v \cos(\omega), & \text{otherwise,}
        \end{cases}
\end{align*}

where $v$ and $\omega$ are local linear and angular velocity published to the mobile robot, respectively. The reward function is composed of two parts: the first part enforces the robot to keep as far as possible away from the obstacles. The second part encourages the robot to move as fast as possible whilst avoiding rotation. The center deviation defines the closeness of the robot to the obstacles and is close to zero if the robot is equally far away from the walls and is increased as soon as the robot moves towards an object. The center deviation is determined by evaluating the range finder readings. The range finder is a sensor with \num{100} line segments covering a \num{270}\si{\degree} \ac{fov} surrounding the agent; its response is the distance to an object intersecting with the ray. The center deviation $c_d$ is calculated as the absolute difference between the sum of right-view and left-view distance measurements of the robot.

If the robot crashes against the wall, the episode ends immediately with a reward of $r_{\text{collision}}=-1$. An episode is considered as solved if the mobile robot navigates without crashing for a total of $T = 2000$ timesteps. The total reward is the sum of the rewards of all steps within an episode. The reward is directly used by the algorithm without clipping or normalization.

\subsection{Policy Learning}
\begin{figure}[t]
\centering
\includegraphics[width=0.85\linewidth]{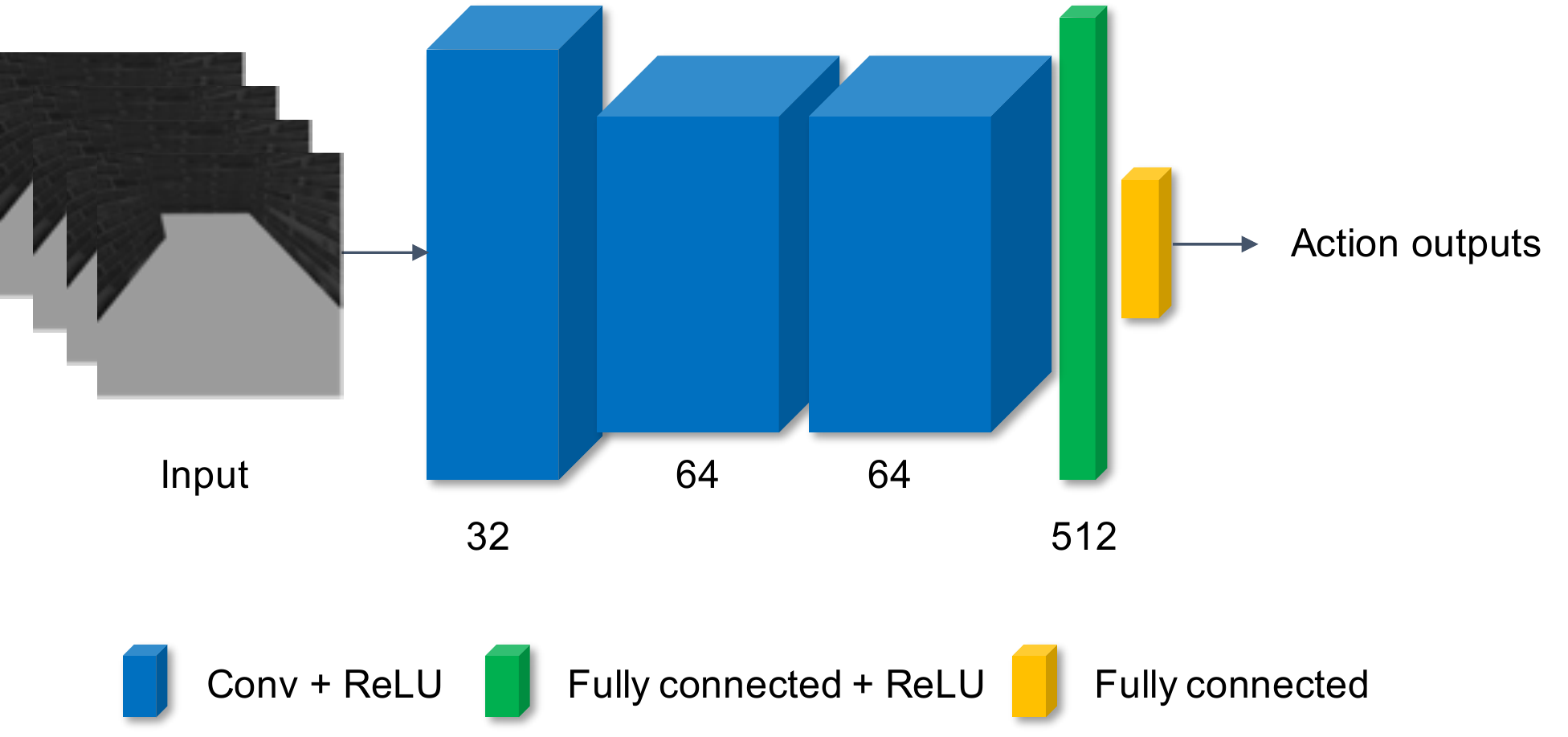}
\caption{Raw sensor information is fed as an input to the network and the output activation is directly used to send control commands to the robot.}
\label{fig:cnn_model}
\end{figure}

For the policy learning, we use the well-known network architecture proposed by Mnih et al.~\cite{Mnih2015}. The architecture is depicted in Figure~\ref{fig:cnn_model}. The network is composed of three convolutional layers having $32$, $64$, and $64$ output channels, respectively with filters of sizes $8 \times 8$, $4 \times 4$, and $3 \times 3$, and strides of $4$, $2$, and $1$. The fully-connected layer has $512$ neurons and is followed by an output layer which differs between the discrete and continuous action space. All hidden layers are followed by rectified linear units (ReLU) as activation functions.

\subsection{Action and Observation Space}
The action space definition varies between the discrete and continuous setting. Each action $a_t$ corresponds to a moving command which is then executed by the mobile robot. 

\textbf{Discrete space.} The space is defined as five predefined actions corresponding to the following commands: \emph{move forward}, \emph{move hard right}, \emph{move soft right}, \emph{move soft left}, and \emph{move hard left}. The linear velocity for the forward direction is set to be \SI{0.3}{\metre\per\second}, while the angular velocity is \SI{0}{\radian\per\second}. The linear velocity for the turning actions is set to be \SI{0.05}{\metre\per\second}, while the angular velocity is $-\frac{\pi}{6}$, $-\frac{\pi}{12}$, $\frac{\pi}{12}$, or $\frac{\pi}{6}$ \si{\radian\per\second} for right and left turns, respectively.

\textbf{Continuous space.} The actions will be sampled initially from a Gaussian distribution with mean $\mu=0$ and standard deviation of $\sigma=1$. For the mapping, the distribution is clipped at $[\mu -3\sigma, \mu +3\sigma]$, since \SI{99.7}{\percent} of the data lie within that band. The linear velocity can take values between the interval $[0.05, 0.3]$ and the angular velocity can take values between $[-\frac{\pi}{6}, \frac{\pi}{6}]$. The policy values from the Gaussian distribution will be linearly mapped to the corresponding action domains.

\textbf{Observation space.} The observation $o_t \in \mathbb{R}^{W \times H \times C}$ (where $W$, $H$, and $C$ are the width, height, and channels of the image) is is an image representation of the scene obtained from an RGB camera mounted on the robot platform. The dimension of the grayscale observation is $W=84$, $H=84$, and $C=1$. The state $s_t$ is composed of four consecutive observations to model temporal dependencies. We assume that the agent interacts with the simulation environment over discrete time steps. At each time step $t$, the agent receives an observation $o_t$ and a scalar reward denoted by $r_t$ from the environment and then transits to the next state $s_{t+1}$. 

\subsection{Training the GAN Architecture}
\begin{figure}[t]
\centering
\includegraphics[width=0.6\linewidth]{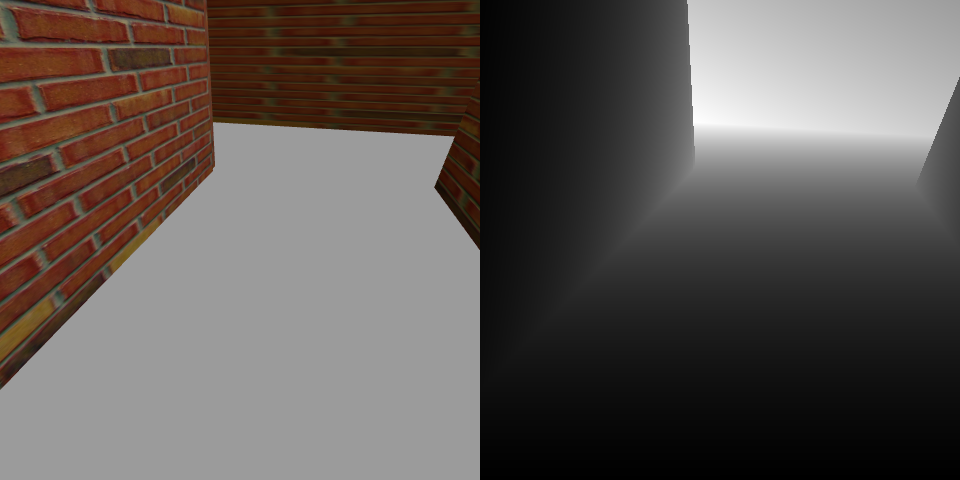}
\caption{A sample image pair used for training the conditional \ac{gan}.}
\label{fig:pix2pix_sample}
\end{figure}

For the training of the paired image-to-image translation, we used the \ac{gan} architecture proposed by Isola et. al~\cite{Isola2017}. The training was done on \num{1313} pairs of pixel-wise corresponding RGB-depth images. An example pair is shown in Figure~\ref{fig:pix2pix_sample}. The image resolution of the images from each domain is resized to $84 \times 84$. The model is trained for a total of \num{200} epochs and the latest model is used for inference during training the \ac{drl} policies. This approach is useful for domain adaptation, where we condition on an input image (RGB image) and generate a corresponding output image (depth image). An advantage of using a \ac{gan} over a simple \ac{cnn} is the fact that this is a generative model and hence not limited to the available training data and can generalize across environments which we validate in our experiments. For more details on the loss and training objective, please see~\cite{Isola2017}.

%% file: sections/experiments.tex
\section{Experimental Results}\label{sec:experiments}

\subsection{Environment Setup}
\begin{figure}[t]
\centering
\includegraphics[width=0.2\textwidth, height=3.5cm]{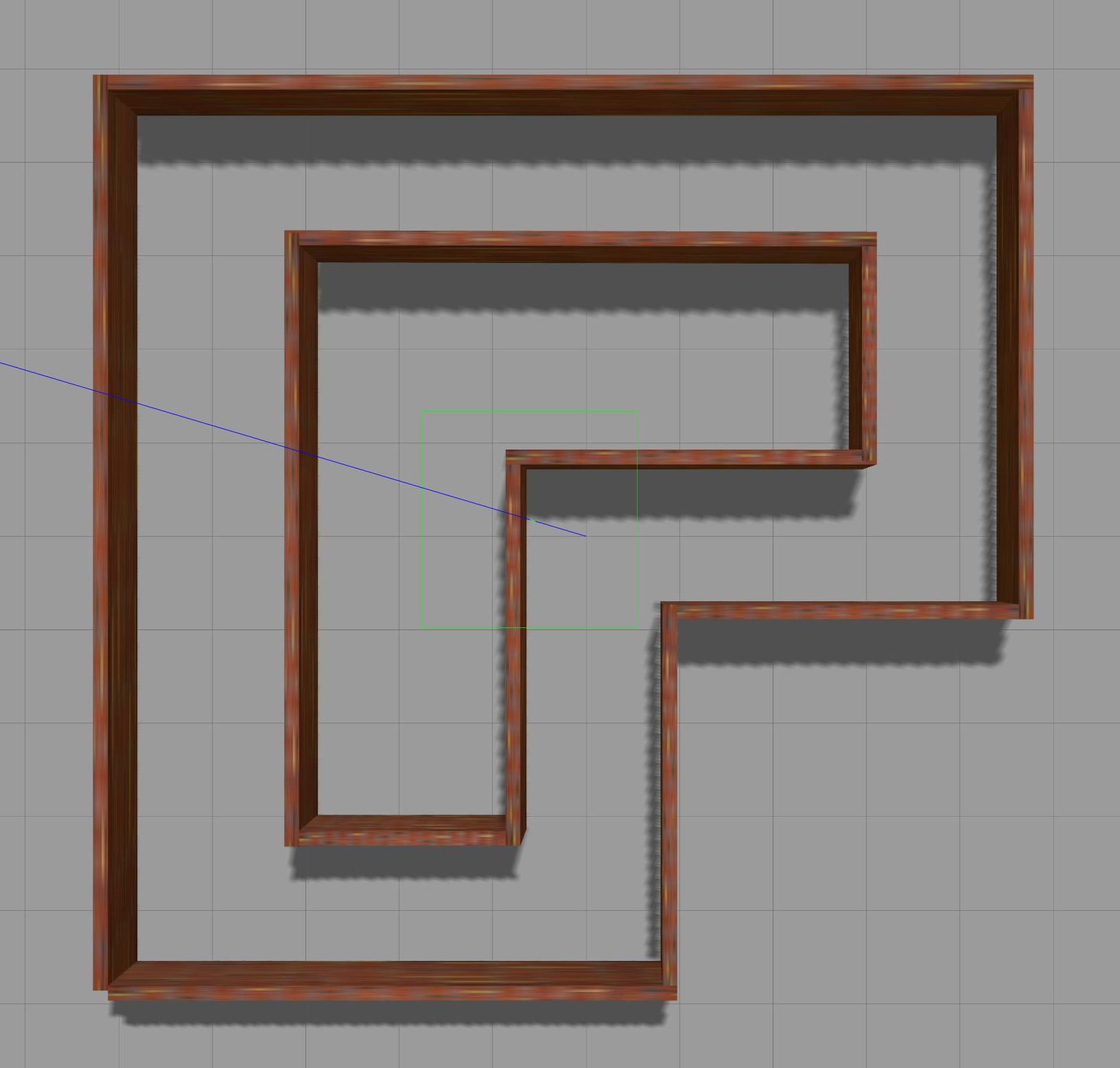}
\hfill
\includegraphics[width=0.2\textwidth, height=3.5cm]{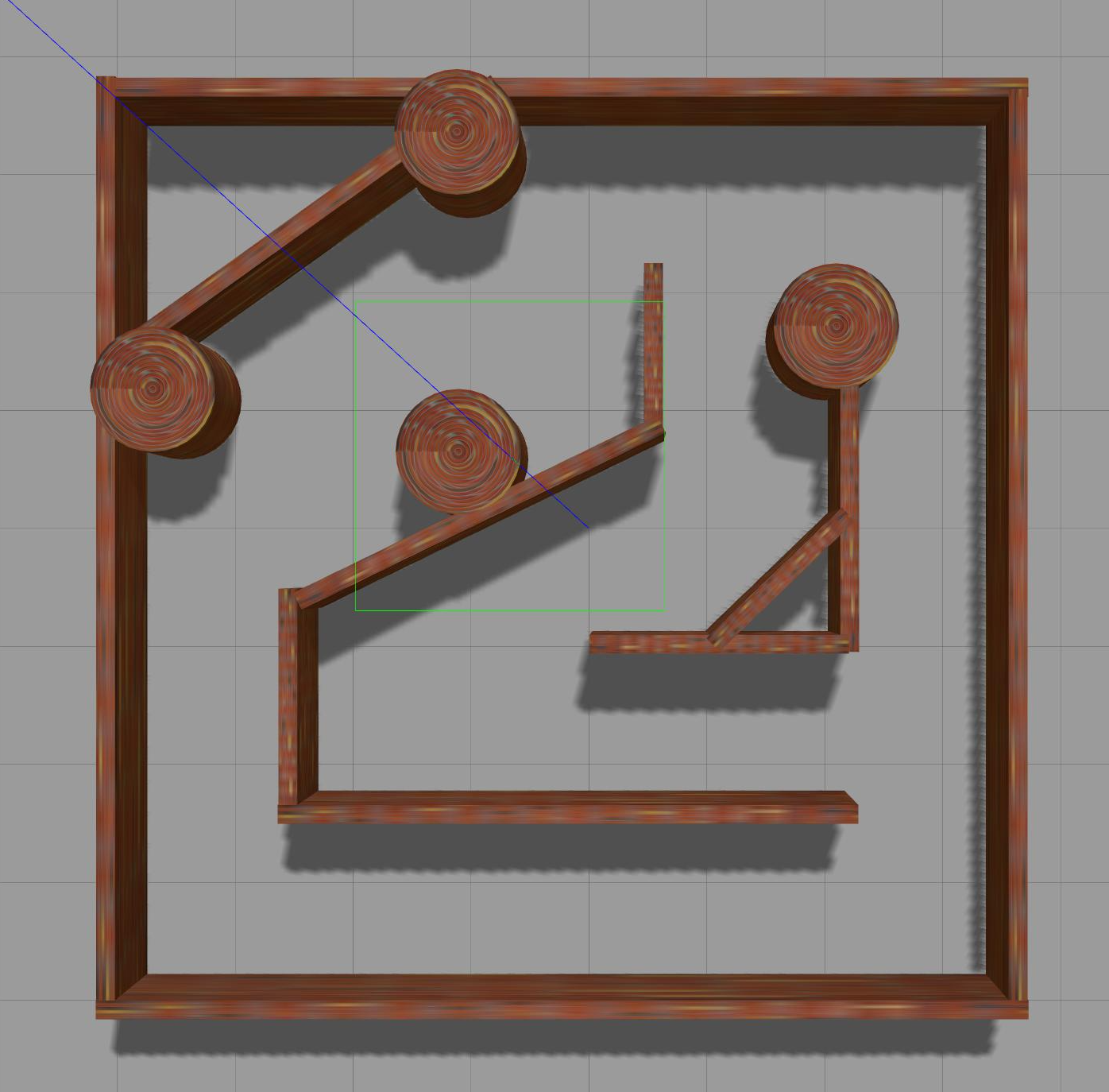}
\caption{The virtual training environments were simulated by Gazebo. We used two different indoor environments with walls around them. The environments are named \texttt{Circuit2-v0}, and \texttt{Maze-v0}, respectively. A TurtleBot2 was used as the robotics platform.}
\label{fig:sim_env}
\end{figure}

We used the gym-gazebo toolkit~\cite{Zamora2016} to evaluate recent \ac{drl} algorithms in different simulation environments. This toolkit is an extension of OpenAI gym~\cite{Brockman2016} for robotics using the Robot Operating System (ROS)~\cite{Quigley2009} and Gazebo~\cite{Koenig2004} that allows for an easy benchmark of different algorithms with the same virtual conditions. For the conducted experiments in this work, two different modified simulation environments from the original project are used; namely \texttt{Circuit2-v0} and \texttt{Maze-v0} (Figure~\ref{fig:sim_env}). As a simulated robotics platform, a TurtleBot2 with a Kobuki base is used. This model comes with a simulated camera with \num{80}\si{\degree} \ac{fov}, $480 \times 480$ resolution, and clipping at $0.05$ and $8.0$ distances. The laser range sensor information is provided by a simple sensor model, which is mounted on top of the robot base. The sensor has a \num{270}\si{\degree} \ac{fov} and outputs sparse range readings ($100$ points). In all our simulation results, each plot shows a \SI{95}{\percent} confidence interval of the mean across $3$ seeds.

\subsection{Training}
For training the models, we use an Adam~\cite{Kingma2014} optimizer and train for a total of $2$ million frames for each environment on an NVIDIA GeForce GTX 1080 GPU. All models are trained with TensorFlow. It is important to know that the laser range finder is only used during training time for determining the rewards and is not required for inference.

In each episode, the mobile robot is randomly spawned in the environment with a random orientation. The angle of the orientation is randomly sampled from an uniform distribution $\phi \sim \mathcal {U}(0, 2\pi)$. To ensure a fair comparison at test time, the agent is spawned at the same location with the same orientation and the reward is averaged over $12$ test episodes in each environment. The average episodic reward is considered as the performance measure of the trained networks. In the following, we present the experimental results.

\subsection{Comparison Between Discrete and Continuous Action Spaces}
\begin{figure}[t]
\centering
\includegraphics[width=0.9\linewidth]{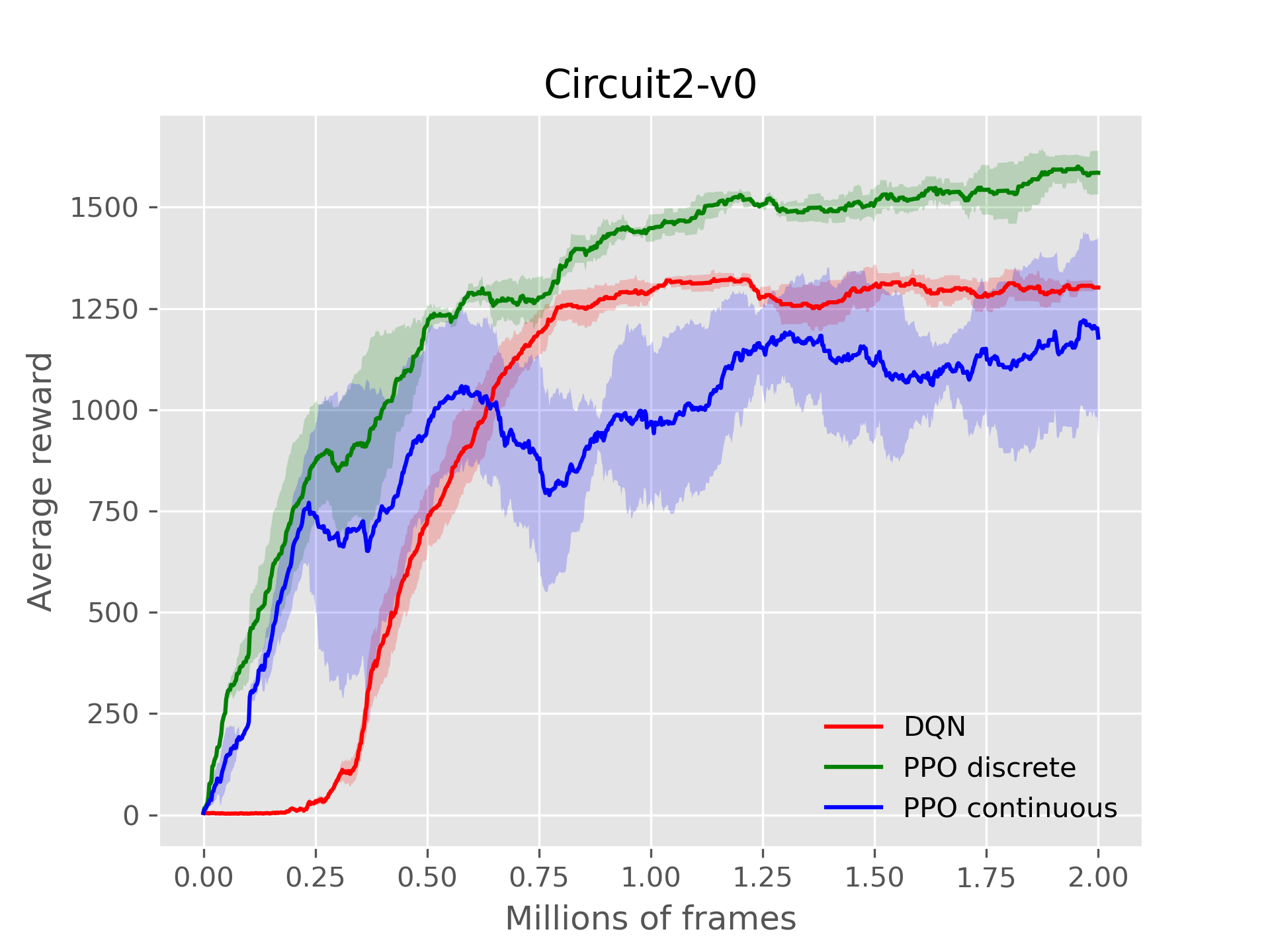}
~
\includegraphics[width=0.9\linewidth]{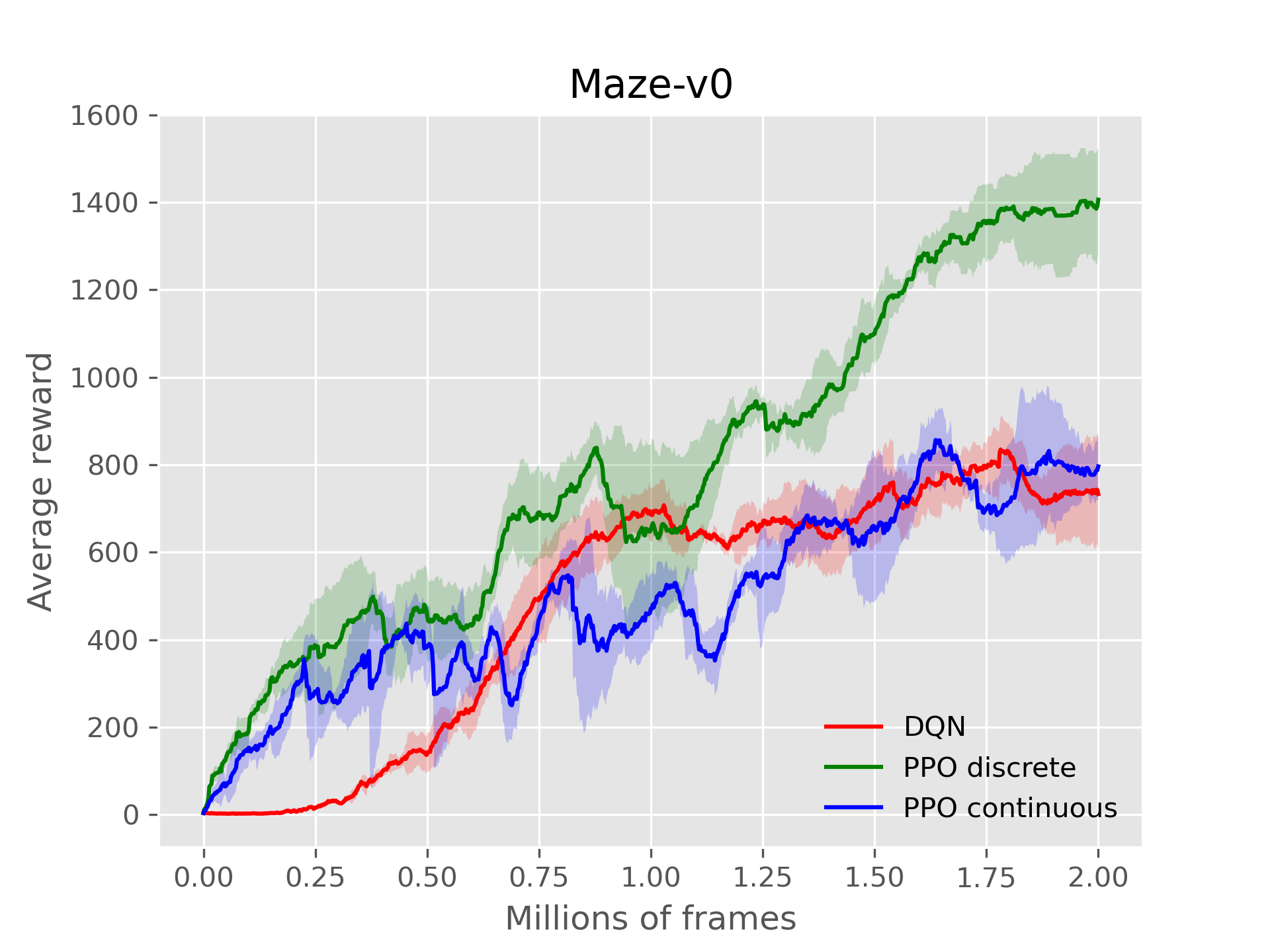}
\caption{The learning curves of DQN (red), PPO discrete (green), and PPO continuous (blue) on the TurtleBot2 robot when trained end-to-end to navigate in an unknown environment. The results depict that all algorithms perform well in the environment and learn a good policy, however, PPO discrete is able to outperform the other methods by a significant margin.}
\label{fig:lr_curves_algos}
\end{figure}

In this experiment, we are particularly interested in finding out the differences of discrete actions compared to continuous actions for visual navigation. Ideally, we want to deal with continuous action spaces for robotic control, since defining discrete action spaces is a cumbersome process and the actions may be limited when performing tricky maneuvers. In Figure~\ref{fig:lr_curves_algos}, the learning curves of DQN (discrete), PPO (discrete), and PPO (continuous) for the \texttt{Circuit2-v0} and \texttt{Maze-v0} environment are shown. The experiments show that PPO discrete is able to outperform both PPO continuous and DQN with discrete action spaces by a wide margin. A viable explanation for the performance differences between PPO discrete and PPO continuous comes from the fact that a relationship between the linear and angular velocity for the continuous domain needs to be learned, whereas prior knowledge about this relation is already available in the discrete domain. Furthermore, in general, the linear and angular velocities are inversely proportional to each other. For example, during turns, the angular velocity is generally high and the linear velocity is low. It is notable to mention that the continuous variant of PPO is able to outperform the discrete DQN method, despite dealing with a much more complicated action space and no prior knowledge about the relationship between linear and angular velocity. Table~\ref{tab:inference_algos} shows the inference results for the different evaluated methods. We see that among the tested methods, PPO with discrete actions is able to achieve the highest scores.  

\begin{table}[t]
\centering
\caption{Inference results for the evaluated \ac{drl} methods for grayscale inputs. We evaluated the performance for $12$ test rollouts, reporting mean and standard deviation.}
\begin{tabular}{lcc}
  \toprule
  & \multicolumn{2}{c}{Average reward}\\
  Method & \texttt{Circuit2-v0} & \texttt{Maze-v0} \\
  \midrule
  DQN & $1327 \pm 15$ & $1016 \pm 80$ \\
  PPO discrete & $\mathbf{1712 \pm 4}$ & $\mathbf{1612 \pm 36}$ \\
  PPO continuous & $1455 \pm 65$ & $1054 \pm 62$ \\
  \bottomrule
\end{tabular}
\label{tab:inference_algos}
\end{table}

\subsection{Comparison Between Varying Discrete Action Spaces}
\begin{figure}[t]
\centering
\includegraphics[width=0.75\linewidth]{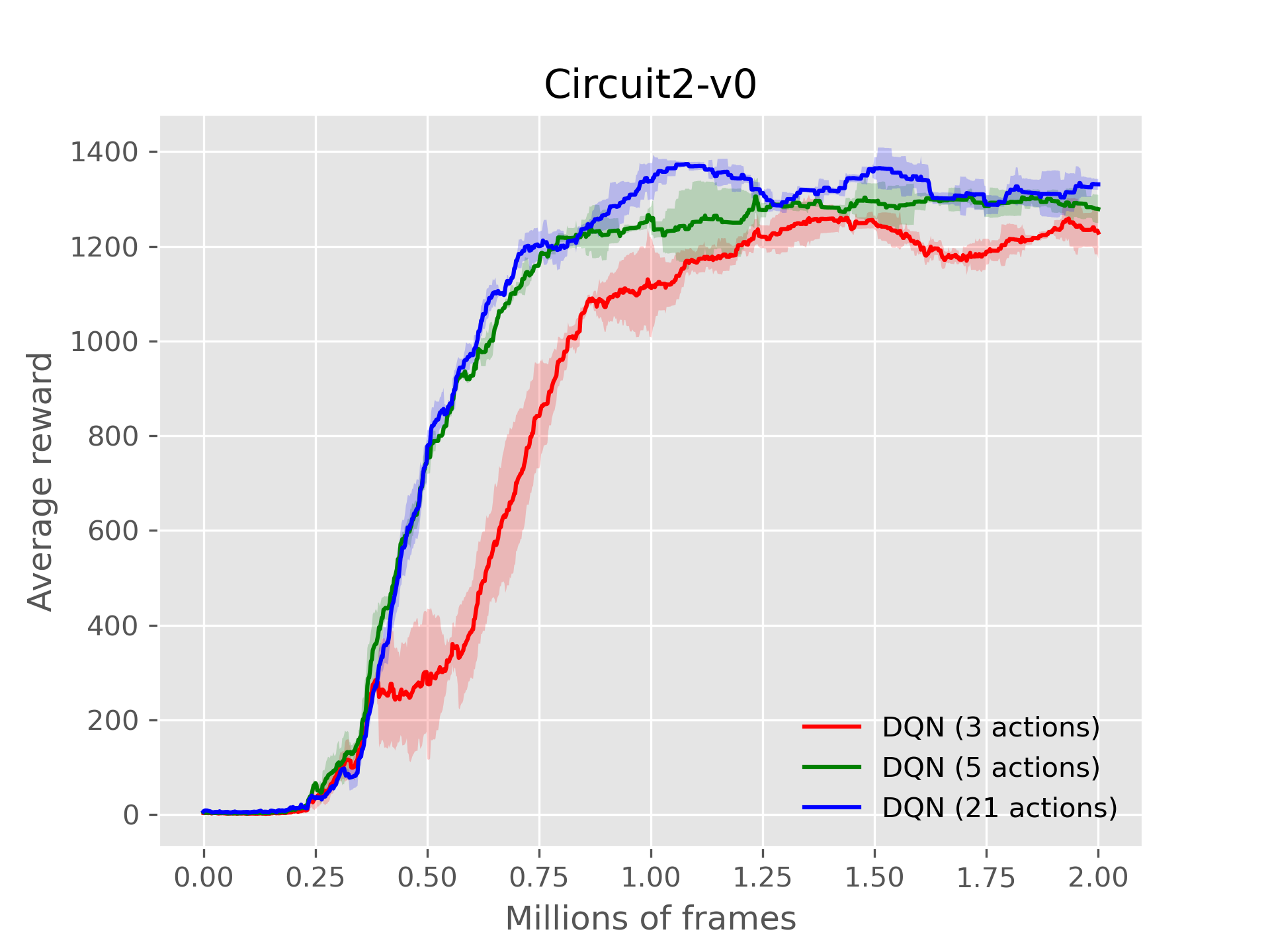}
~
\includegraphics[width=0.75\linewidth]{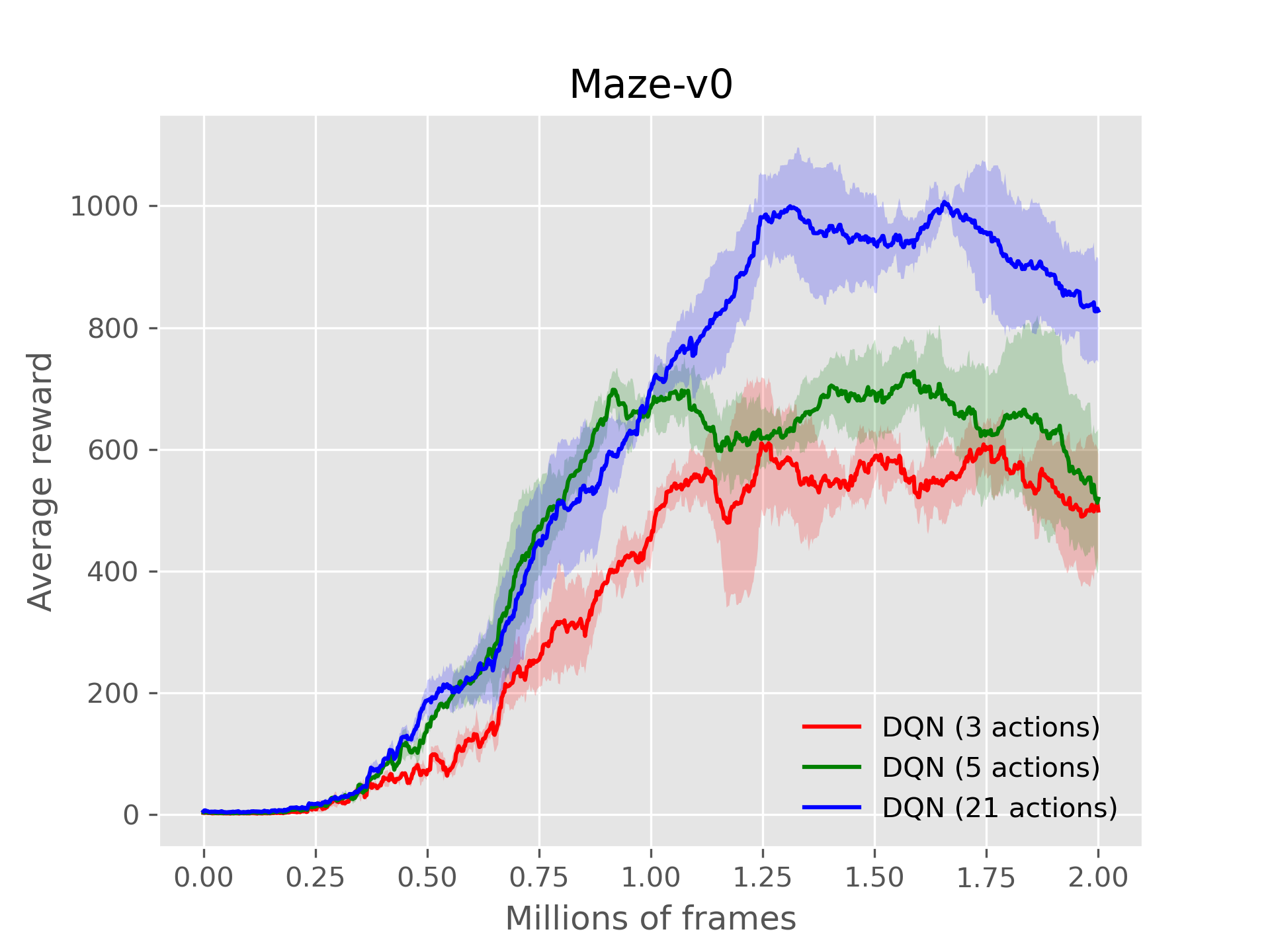}
\caption{The learning curves of DQN (3 actions), DQN (5 actions), and DQN (21 actions).}
\label{fig:lr_curves_different_actions}
\end{figure}

In this experiment, the impact of the number of actions on the learning performance is analyzed. The linear and the angular velocity for the \emph{move forward} action is the same as before and only the angular velocities of the turning actions are varied. Table~\ref{tab:action_spaces_many} shows the corresponding angular velocities for the different action spaces. We are comparing $3$ different numbers of actions. In Figure~\ref{fig:lr_curves_different_actions}, the learning curves of the different variants of DQN are shown. The experiments show that a larger number of actions results in a higher average reward in both evaluated environments. Moreover, the experiments show that more fine-grained action spaces are particularly helpful in environment \texttt{Maze-v0}, where the agent has to perform more complicated maneuvers as compared to the ones in environment \texttt{Circuit2-v0}.

\begin{table}[t]
\centering
\caption{Angular velocities for different action spaces. Velocities sorted from rightmost to leftmost angular direction.}
\renewcommand{\arraystretch}{1.3}
\begin{tabular}{lc}
  \toprule
  Number of actions & Angular velocity (in \si{\radian\per\second}) \\
  \midrule
  3 & $-0.3$, $0$, $0.3$ \\
  5 & $-\frac{\pi}{6}$, $-\frac{\pi}{12}$, $0$, $\frac{\pi}{12}$, $\frac{\pi}{6}$ \\
  21 & $-\frac{\pi}{6} + \frac{\pi n}{60} \text{ for } n=0,\dotsc,20$ \\
  \bottomrule
\end{tabular}
\label{tab:action_spaces_many}
\end{table}

\subsection{Comparison Between Sensor Modalities}
\begin{figure}[ht]
    \centering
    \begin{subfigure}[c]{0.15\textwidth}
        \centering
        \includegraphics[height=.8in]{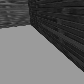}
        \caption{Grayscale}
    \end{subfigure}%
    \hfill
    \begin{subfigure}[c]{0.15\textwidth}
        \centering
        \includegraphics[height=.8in]{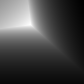}
        \caption{Depth}
    \end{subfigure}%
    \hfill
    \begin{subfigure}[c]{0.15\textwidth}
        \centering
        \includegraphics[height=.8in]{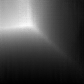}
        \caption{Predicted depth}
    \end{subfigure}%
    \caption{Examples of the different input modalities. From left to right: monocular grayscale image, ground-truth depth image, and predicted depth image by the conditional \ac{gan} approach.}
    \label{fig:sensor_inputs}
\end{figure}

In the previous section, all methods were trained on a single visual input modality. In this experiment, we compare agents equipped with different sensor modalities to investigate the impact of fused sensory inputs. The different sensor modalities are depicted in Figure~\ref{fig:sensor_inputs}.

\begin{figure}[t]
\centering
\includegraphics[width=0.75\linewidth]{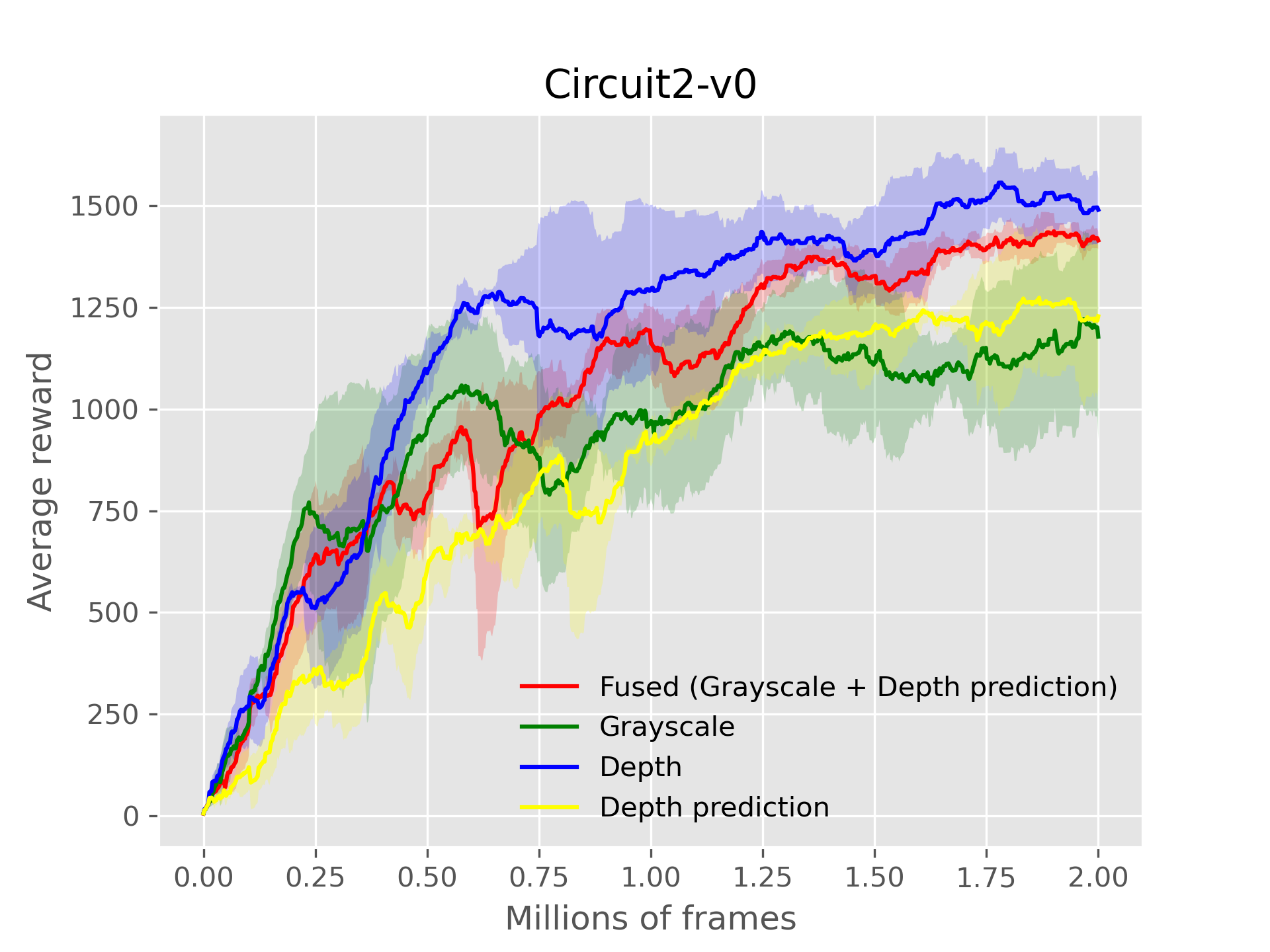}
~
\includegraphics[width=0.75\linewidth]{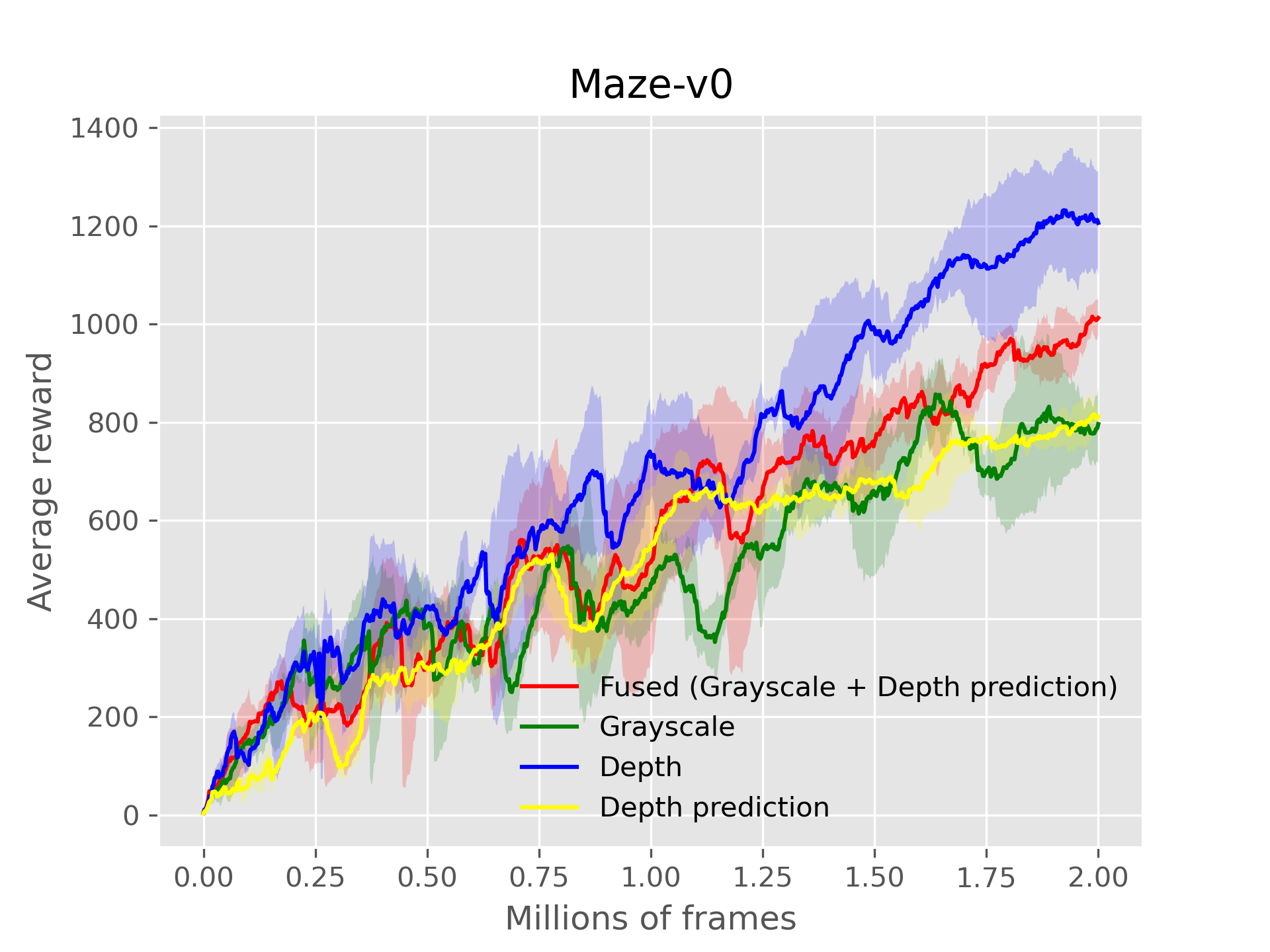}
\caption{The learning curves of Fused (red), Grayscale (green), Depth (blue), and Depth prediction (yellow) on the TurtleBot2 robot when trained end-to-end to navigate in an unknown environment. The results depict that all sensor modalities perform well in the environment and learn a good policy, however, the depth sensor is able to outperform the monocular camera by a significant margin. By fusing predicted depth images with grayscale images, we can increase the learning performance.}
\label{fig:lr_curves_sensors}
\end{figure}

We propose to fuse multiple sources to obtain improved learning performance. The idea is to combine the grayscale images with depth images which are predicted by a conditional \ac{gan}. This is done by concatenating a grayscale image with the predicted depth image and using it as an observation. This sensor combination is denoted as Fused. In Figure~\ref{fig:lr_curves_sensors}, the learning curves with Grayscale, Depth, Depth prediction, and Fused sensor inputs for the \texttt{Circuit2-v0} and \texttt{Maze-v0} environment are shown. All sensor configurations are trained with the PPO continuous method. From the red learning curve, we can see that the advantage of using the Fused approach increases the average reward compared to using the grayscale input only. From the figure, it can be seen that the depth image is also advantageous in terms of learning performance compared to the raw grayscale input. This result is consistent with earlier research results of Peasley~\cite{Peasley2013} which shows that depth information is vital for tasks like exploration and obstacle avoidance.

Table~\ref{tab:inference_fused} shows the inference results for the different evaluated sensor setups. We see that among the tested sensor modalities, the depth image achieves the highest average reward. The difference in the average reward between using depth images or grayscale images as input is significant. However, using our proposed Fused sensor setup, we can achieve almost similar results as with using ground truth depth images.

\begin{table}[ht]
\centering
\caption{Inference results for the evaluated \ac{drl} methods for different sensor modalities. We evaluated the performance for $12$ test rollouts, reporting mean and standard deviation.}
\begin{tabular}{lcc}
  \toprule
  & \multicolumn{2}{c}{Average reward}\\
  Sensor modality & \texttt{Circuit2-v0} & \texttt{Maze-v0} \\
  \midrule
  Fused & $1619 \pm 5$ &$1411 \pm 83$ \\
  Grayscale & $1455 \pm 65$ & $1054 \pm 62$ \\
  Depth & $\mathbf{1627 \pm 9}$ & $\mathbf{1474 \pm 61}$ \\
  Depth prediction & $1544 \pm 43$ & $896 \pm 22$ \\
  \bottomrule
\end{tabular}
\label{tab:inference_fused}
\end{table}

%% file: sections/conclusion.tex
\section{Conclusion}\label{sec:conclusion}
In this paper, we have presented a \ac{drl} approach for vision-based obstacle avoidance with a mobile robot from raw sensory data. Our method solely relies on a monocular camera for perception, which enables low-cost, lightweight, and low-power-consuming hardware solutions. We validated our approach on several simulated experiments, moreover, we analyzed the applicability of two different kinds of algorithms, namely PPO and DQN. We showed that control policies for discrete and continuous action spaces can be learned in an end-to-end manner. Our experiments show that these algorithms can learn to navigate in simple maze-like environments without prior knowledge of the environment. Furthermore, we showed how the average reward can be increased by additionally fusing predicted depth images by means of a \ac{gan}. Moreover, the impact of the number of actions on the learning behavior is analyzed. This result indicates a promising direction to continue research on \ac{drl} for mapless navigation. In future work, we intend to work on transferring the policies trained in simulation to realistic environments. Furthermore, another interesting direction would be to use imitation learning techniques to solve the sample-inefficiency problem and potentially make the training process faster and safer.